\documentclass[runningheads]{llncs}
\usepackage[T1]{fontenc}
\usepackage{cite}
\usepackage{hyperref}
\usepackage{url}
\usepackage{graphicx}
\usepackage{subcaption}
\usepackage{booktabs}
\usepackage{tabularx}
\usepackage{enumitem}
\usepackage{wrapfig}
\usepackage{multirow}
\usepackage{tcolorbox}
\usepackage{bm}
\usepackage{amsmath}
\usepackage{pifont}
\usepackage[table]{xcolor}
\renewcommand{\thefootnote}{\arabic{footnote}}

\definecolor{cFirebrick}{RGB}{178, 34, 34}
\definecolor{cPaleTurquoise}{RGB}{175, 238, 238}
\definecolor{cSienna}{RGB}{160, 82, 45}
\definecolor{cSkyBlue}{RGB}{135, 206, 235}
\definecolor{cOliveDrab}{RGB}{107, 142, 35}
\definecolor{cThistle}{RGB}{216, 191, 216}
\definecolor{cDarkOrchid}{RGB}{153, 50, 204}
\definecolor{cPaleGreen}{RGB}{152, 251, 152}
\definecolor{cSteelBlue}{RGB}{70, 130, 180}
\definecolor{cSandyBrown}{RGB}{244, 164, 96}
\definecolor{cMediumSeaGreen}{RGB}{60, 179, 113}
\definecolor{cLightCoral}{RGB}{240, 128, 128}
\definecolor{cChocolate}{RGB}{210, 105, 30}
\definecolor{cLightGreen}{RGB}{144, 238, 144}
\definecolor{cKhaki}{RGB}{240, 230, 140}
\definecolor{cOrchid}{RGB}{218, 112, 214}
\definecolor{cDarkKhaki}{RGB}{189, 183, 107}
\definecolor{cPlum}{RGB}{221, 160, 221}
\definecolor{cBrown}{RGB}{165, 42, 42}
\definecolor{cIndigo}{RGB}{75, 0, 130}
\definecolor{cCadetBlue}{RGB}{95, 158, 160}
\definecolor{cPaleGoldenrod}{RGB}{238, 232, 170}
\definecolor{cMediumTurquoise}{RGB}{72, 209, 204}
\definecolor{cGrey}{RGB}{128, 128, 128}
\definecolor{cSlateGrey}{RGB}{112, 128, 144}
\definecolor{cMediumSlateBlue}{RGB}{123, 104, 238}
\definecolor{cWheat}{RGB}{245, 222, 179}
\definecolor{cLightPink}{RGB}{255, 182, 193}
\definecolor{cDarkSalmon}{RGB}{233, 150, 122}
\definecolor{cTan}{RGB}{210, 180, 140}
\definecolor{cMediumPurple}{RGB}{147, 112, 219}
\definecolor{cIndianRed}{RGB}{205, 92, 92}
\definecolor{cAquamarine}{RGB}{127, 255, 212}
\definecolor{cSlateGray}{RGB}{112, 128, 144}
\definecolor{cBlanchedAlmond}{RGB}{255, 235, 205}
\definecolor{cLightSteelBlue}{RGB}{176, 196, 222}
\begin{document}
\title{HueManity: Probing Fine-Grained Visual Perception in MLLMs}
%
%\titlerunning{Abbreviated paper title}
% If the paper title is too long for the running head, you can set
% an abbreviated paper title here
%
\author{Rynaa Grover\inst{1}\thanks{Equal Contributions} \and
Jayant Sravan Tamarapalli\inst{1}\textsuperscript{\thefootnote} \and
Sahiti Yerramilli\inst{1}\textsuperscript{\thefootnote} \and
Nilay Pande\inst{2}\textsuperscript{\thefootnote}}
\authorrunning{R. Grover, J.S. Tamarapalli, S. Yerramilli, and N. Pande.}
% First names are abbreviated in the running head.
% If there are more than two authors, 'et al.' is used.
%
\institute{Google, Mountain View, CA, USA \\ \email{\{rynaa,jayantsravan,sahitiy\}@google.com}\\ \and
Waymo, Mountain View, CA, USA \\ \email{nilayp@waymo.com}\\}
\maketitle              % typeset the header of the contribution
\begin{figure*}[!h]
    \centering
    \includegraphics[width=1\linewidth]{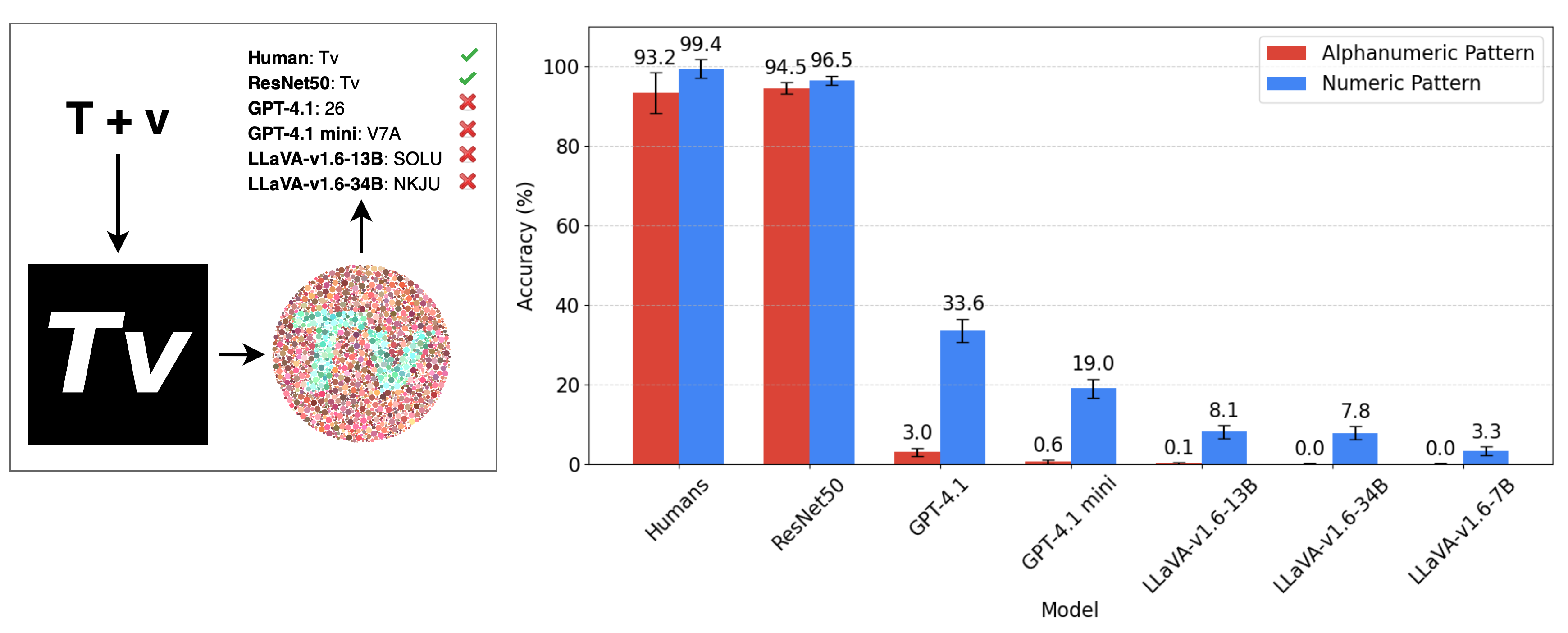}
    \caption{We present \textbf{HueManity} - an automated scalable benchmark for evaluation of fine-grained visual perception in MLLMs. The pipeline (left) embeds characters within challenging Ishihara-style image patterns, while ensuring human readability of the generated images. Experiments reveal (right) that humans and a fine-tuned ResNet50 baseline significantly outperform top-5 leading closed-source and open-source MLLMs, exposing a critical lack of fine-grained understanding.}
    \label{fig:overview}
\end{figure*}

\begin{abstract}
% Multimodal Large Language Models (MLLMs) demonstrate strong high-level visual reasoning, yet their foundational understanding of nuanced perceptual details is often overlooked by existing evaluations. To address this, we introduce HueManity, a benchmark designed to assess visual perception in MLLMs. HueManity comprises 83,850 Ishihara-style images with embedded alphanumeric strings, challenging models on pattern recognition - a fundamental aspect of visual understanding. Our evaluation of nine MLLMs on HueManity demonstrates a significant performance deficit: the best-performing model achieved only 33.6\% accuracy on an `easy' numeric task and 3\% on a `hard' alphanumeric task. This starkly contrasts with human (99.38\% numeric, 93.25\% alphanumeric)\footnote{Human were evaluated on a 100-image representative subsets, sampled from the model evaluation sets for each task.} and fine-tuned ResNet50 (96.5\% numeric, 94.5\% alphanumeric) performance. These findings uncover a critical gap in MLLMs' fine-grained visual understanding, a limitation not apparent through conventional tasks that focus on high-level semantic assessments.
% HueManity offers a new paradigm for evaluating this specific type of model understanding. We will open-source the dataset and code to foster further research into this critical area.

Recent Multimodal Large Language Models (MLLMs) demonstrate strong high-level visual reasoning on tasks such as visual question answering and image captioning. Yet existing benchmarks largely overlook their ability to capture fine-grained perceptual details. As MLLMs are increasingly deployed in safety and reliability critical settings, perceptual acuity becomes essential. We present HueManity, a scalable automated benchmark for assessing fine-grained visual perception in MLLMs. HueManity comprises 83,850 Ishihara-style images embedding alphanumeric strings, designed to evaluate pattern recognition, a core aspect of visual understanding. Our evaluation of nine state-of-the-art MLLMs uncovers a striking performance deficit: the strongest model achieved only 33.6\% accuracy on a simple numeric task and 3\% on a harder alphanumeric task, compared to near-ceiling performance from humans (99.38\%, 93.25\%) and a fine-tuned ResNet-50 (96.5\%, 94.5\%). These findings expose a critical weakness in MLLMs’ perceptual grounding, one that remains obscured by conventional benchmarks emphasizing high-level semantics.

\textbf{Code:} \href{https://github.com/rynaa/huemanity}{https://github.com/rynaa/huemanity}

\textbf{Dataset:} \href{https://huggingface.co/datasets/Jayant-Sravan/HueManity}{https://huggingface.co/datasets/Jayant-Sravan/HueManity}

\keywords{Visual Perception \and Multimodal Large Language Models \and Pattern Recognition \and Robustness}
\end{abstract}

\section{Introduction} \label{sec:introduction}
The trajectory of Multimodal Large Language Models (MLLMs) \cite{comanici2025gemini25pushingfrontier, gpt4technicalreport, qwenvl, blip2, multimodalgpt, llava, anthropic2025claude} has been marked by impressive advancements, demonstrating sophisticated capabilities in bridging visual and textual information. Their skills extend beyond simple image labeling~\cite{russakovsky2015imagenetlargescalevisual, 6296535}, enabling complex tasks like generating detailed image descriptions \cite{dong2024dreamllmsynergisticmultimodalcomprehension,  fu2024guidinginstructionbasedimageediting}, answering intricate visual questions requiring inference about relationships and activities \cite{weng2025learningcompresscontextsefficient, chen-etal-2025-seeing, kuang2024naturallanguageunderstandinginference}, and participating in nuanced dialogue about visual content \cite{cao-etal-2024-visdiahalbench}. This success is largely attributed to pre-training on vast, web-scale image-text datasets, which has cultivated a powerful ability to map high-level semantic concepts between modalities~\cite{jia2021scalingvisualvisionlanguagerepresentation, radford2021learningtransferablevisualmodels, schuhmann2022laion5bopenlargescaledataset, alayrac2022flamingovisuallanguagemodel, qi2020imagebertcrossmodalpretraininglargescale, zhai2022litzeroshottransferlockedimage, pham2023combinedscalingzeroshottransfer}.

However, this focus on semantic understanding, both in training and in predominant evaluation paradigms~\cite{touchstone, seedbench1, seedbench2, lvlmehub, lamm, llava}, has created a critical blind spot: the models' foundational perceptual acuity remains largely unprobed. Human vision is not merely a semantic engine, it is fundamentally a system for perceptual organization, excelling at extracting coherent signals from visually cluttered environments. This ability to discern patterns from subtle, low-level cues — like color, luminance, and texture — is a prerequisite for virtually all higher-level visual reasoning. 
% This is a skill honed by millennia of evolution; before our ancestors could reason `that is a tiger,' they first had to perceptually distinguish the subtle pattern of its stripes from the chaotic background of jungle leaves.
%Before we can identify a ``cat in a tree," we must first be able to perceptually group the pixels corresponding to the cat and distinguish them from the noisy background of leaves and branches.
%As foundation models increasingly shape critical applications, ensuring high visual acuity in MLLMs is essential—for instance, an autonomous vehicle relying on them could misinterpret a traffic light in rainy weather, leading to life-threatening collisions 

Existing benchmarks have predominantly centered on the conceptual capabilities of MLLMs, leaving their resilience to perceptual challenges like pattern recognition and feature differentiation in cluttered scenes largely unevaluated. This paper introduces \textbf{HueManity}, a benchmark specifically designed to probe this gap. Our methodology is inspired by Ishihara plates~\cite{clark1924ishihara}, a classic tool from human ophthalmology designed to isolate a specific perceptual skill: figure-ground segregation based on subtle color cues. It is crucial to clarify that HueManity does not aim to diagnose `color blindness' in MLLMs. Instead, our Ishihara-style stimuli, created using controlled generation techniques, test an MLLM's fundamental ability to identify embedded alphanumeric characters by their subtle color and luminance contrasts within visually cluttered dot patterns.

The ability to parse characters from our Ishihara-style plates is a direct proxy for an MLLM's capacity to handle real-world visual challenges characterized by clutter, partial occlusions, and variable lighting. An MLLM that cannot distinguish a number embedded in a field of dots may likewise struggle to perform optical character recognition (OCR) on text seen through a rainy window or accurately transcribe text from a crumpled, faded receipt. Therefore, HueManity serves not merely as a test of pattern recognition, but as a diagnostic tool to probe the limitations that prevent MLLMs from achieving robust, human-like visual intelligence. Our findings reveal that this novel evaluation exposes a profound deficit in current models, a result with significant implications for deploying MLLMs in domains requiring human-like perceptual understanding, from autonomous systems to medical imaging \cite{yang2025illusions}.

To address this identified gap and facilitate further research in this domain, this paper makes the following contributions:
\begin{enumerate}
    \item \textbf{We introduce HueManity, a new large-scale benchmark ($\text{83,850}$ images)} featuring Ishihara-inspired alphanumeric stimuli. The benchmark utilizes a principled design with $\text{25}$ carefully curated color pairs, selected using $\text{CIEDE2000}$ ($\Delta E_{2000}$) metrics and manual verification, ensuring both systematic challenge and fairness for human comparison.
    \item \textbf{We conduct a comprehensive evaluation of nine state-of-the-art MLLMs}, revealing a significant performance gap when compared to strong human and fine-tuned ResNet50 baselines. This suggests MLLM limitations are architectural rather than the task being intractable
    \item \textbf{We release open-source code for generating challenging Ishihara-style perceptual stimuli}, enabling reproducible research and community-driven extensions.
\end{enumerate}

\section{Related Works} \label{sec:related}
\subsection{Multimodal Models}
With the remarkable advancements of Large Language Models (LLMs), recent research has extended their capabilities to multimodal domains by integrating visual information, giving rise to Multimodal Large Language Models (MLLMs) \cite{comanici2025gemini25pushingfrontier, gpt4technicalreport, qwenvl, blip2, multimodalgpt, llava}. These models typically align visual features from pre-trained image encoders with LLMs via modality adaptation layers. Early works like BLIP-2 \cite{blip2} pioneered this architecture by first pre-training on image-text datasets and fine-tuning on task-specific benchmarks such as Visual Question Answering (VQA). Subsequent models like LLaVA \cite{llava} advanced this approach by leveraging synthetic instruction-following data in VQA formats, significantly improving instruction tuning performance. More recent efforts have expanded into video understanding and even image generation \cite{baldridge2024imagen, saharia2022photorealistic}, showcasing the versatility of MLLMs across modalities. However, this celebrated success in visual tasks often appears reliant on their powerful language capabilities for reasoning and interpretation, potentially overshadowing the need to scrutinize their fundamental perception skills. Addressing this gap, HueManity is a benchmark designed specifically to isolate and probe these visual abilities.

\subsection{MLLM Evaluation}
While MLLMs excel at global image understanding, they often struggle with fine-grained tasks requiring precise recognition and localization \cite{survey, seedbench1}. Existing benchmarks designed to probe these limitations face three primary methodological challenges:

\textbf{Semantic Entanglement}:In many perception benchmarks, like Eyes Wide Shut \cite{tong2024eyes}, MERLIM \cite{villa2025magicmerlimmultimodalevaluation}), perception is often ``entangled'' with reasoning. Questions focusing on semantic concepts like ``left/right'' or ``yellow animal'' allow models to rely on language priors rather than pure visual input.

\textbf{Evaluation}: Early benchmarks like LLaVA-Bench \cite{llava}, and TouchStone \cite{touchstone} rely on GPT-based judges. However, LLM-based evaluation is unreliable for fine-grained perception since the best models are also not very adept. Other efforts, such as LVLM-eHub \cite{lvlmehub}, depend on expensive human annotators, limiting scalability.

\textbf{Data Integrity and Scale}: To improve reliability, MME \cite{mme}, MMBench \cite{mmbench}, and SEED-Bench \cite{seedbench1} adopted objective, multiple-choice formats. However, these often draw from existing VQA datasets, raising concerns about data contamination. While ZeroBench \cite{zerobench} offers increased difficulty, its utility for fine-grained perception is limited by its broad focus and small scale. Similarly, illusion-based benchmarks \cite{hallusionbench, illusionvqa} are often constrained by small dataset sizes and a reliance on common internet images.

HueManity departs from these approaches by \textbf{explicitly disentangling perception from semantics}. Unlike benchmarks where linguistic context could help guess an answer, HueManity provides no semantic cues to predict a hidden digit like `26'. The model must perceive the underlying dot patterns directly, making it impossible to rely on language priors. Through procedural generation and an exact-match evaluation framework, HueManity offers a novel, scalable, and entirely objective methodology for evaluating a core perceptual capability that remains a challenge for state-of-the-art MLLMs.

\begin{table}[h]
  \caption{Comparison of MLLM evaluation benchmarks across key methodological attributes. Novelty refers to the use of non-internet, generated images.} 
  \label{related-works-comparison}
  \centering
  % Use tabularx to make the table a specific width and wrap text
  \begin{tabularx}{1\textwidth}{@{} l *{5}{>{\centering\arraybackslash}X} @{}}
    \toprule
    \textbf{Benchmark} & \textbf{Data Size} & \textbf{Automatic Annotation} & \textbf{Novel Images} & \textbf{Answer Form} & \textbf{Automated Evaluation} \\
    \midrule % Use \midrule to separate the header from the body
    LVLM-eHub & - & \ding{51} & \ding{55} & Free-form & \ding{55} \\ 
    MMBench & 3,217 & \ding{55} & \ding{55} & Multi-choice & \ding{51}\\
    SEED-Bench2 & 24,371 & \ding{55} & \ding{55} & Multi-choice & \ding{51}\\
    Blink & 3,807 & \ding{55} & \ding{55} & Multi-choice & \ding{51}\\
    ZeroBench & 100 & \ding{55} & \ding{51} & Free-form & \ding{51}\\
    \textbf{HueManity} & 83,850 & \ding{51} & \ding{51} & Exact Match & \ding{51} \\
    \bottomrule
  \end{tabularx}
\end{table}

\section{Data Creation} \label{sec:data-creation}
The HueManity benchmark is built upon a dataset of 83,850 images, each featuring a two-character alphanumeric string rendered as an Ishihara-style dot pattern, along with its corresponding ground truth label and generation parameters. Our generation process involves creating a base text mask, selecting a color palette, and rendering the final dot pattern. Figure \ref{fig:overview} illustrates the core stages of this pipeline.

% \begin{itemize}
%     \item \textit{Number Recognition Set (Easier Task)}: This subset contains images with two-digit numeric strings (e.g., 17, 83, 65). We exclude leading-zero numbers ('00'-'09') to prevent ambiguity between single and double-digit interpretations.
%     \item \textit{Alphanumeric Recognition Set (Harder Task)}: This subset includes two-character strings (e.g., A7, 9b, XG) formed from lowercase letters (a-z), uppercase letters (A-Z), and digits (0-9). We excluded visually ambiguous characters (`l', `I', `J', `O') from the character space.
% \end{itemize}

% Additionally, to isolate the challenge of perceptual grouping from basic character recognition, we evaluated all models under two distinct visual conditions for each image:
% \begin{itemize}
%     \item \textit{Ishihara Pattern:} The primary task using the Ishihara-style dot pattern images.
%     \item \textit{Text Masks:} A control task using the binary text mask (Figure~\ref{fig:number-mask}) used in the creation of Ishihara Pattern images. This condition establishes a baseline for each model's fundamental OCR capability, helping to disentangle it from performance on the more perceptually complex patterns.
% \end{itemize}

\textbf{Text Mask Generator.}
The first step is to create a 900x900 pixel binary mask for each two-character string. We use the Pygame library to render the string in white on a black background. To ensure the characters are thick enough for dot-based rendering and remain clearly legible, we use the DejaVu Sans font, styled in bold and italic at a size of 550 (Figure \ref{fig:overview}).

\textbf{Color Pairs Selection}
To generate the patterns, we meticulously selected 25 distinct foreground-background color pairs. This selection process involved a multi-stage procedure combining quantitative analysis using the CIEDE2000 color-difference formula \cite{article} with extensive manual verification. This ensures that the color pairs are balanced for perceptual difficulty while remaining legible to humans (see Appendix D for details).

\begin{figure}
    \centering
    \includegraphics[width=0.7\linewidth]{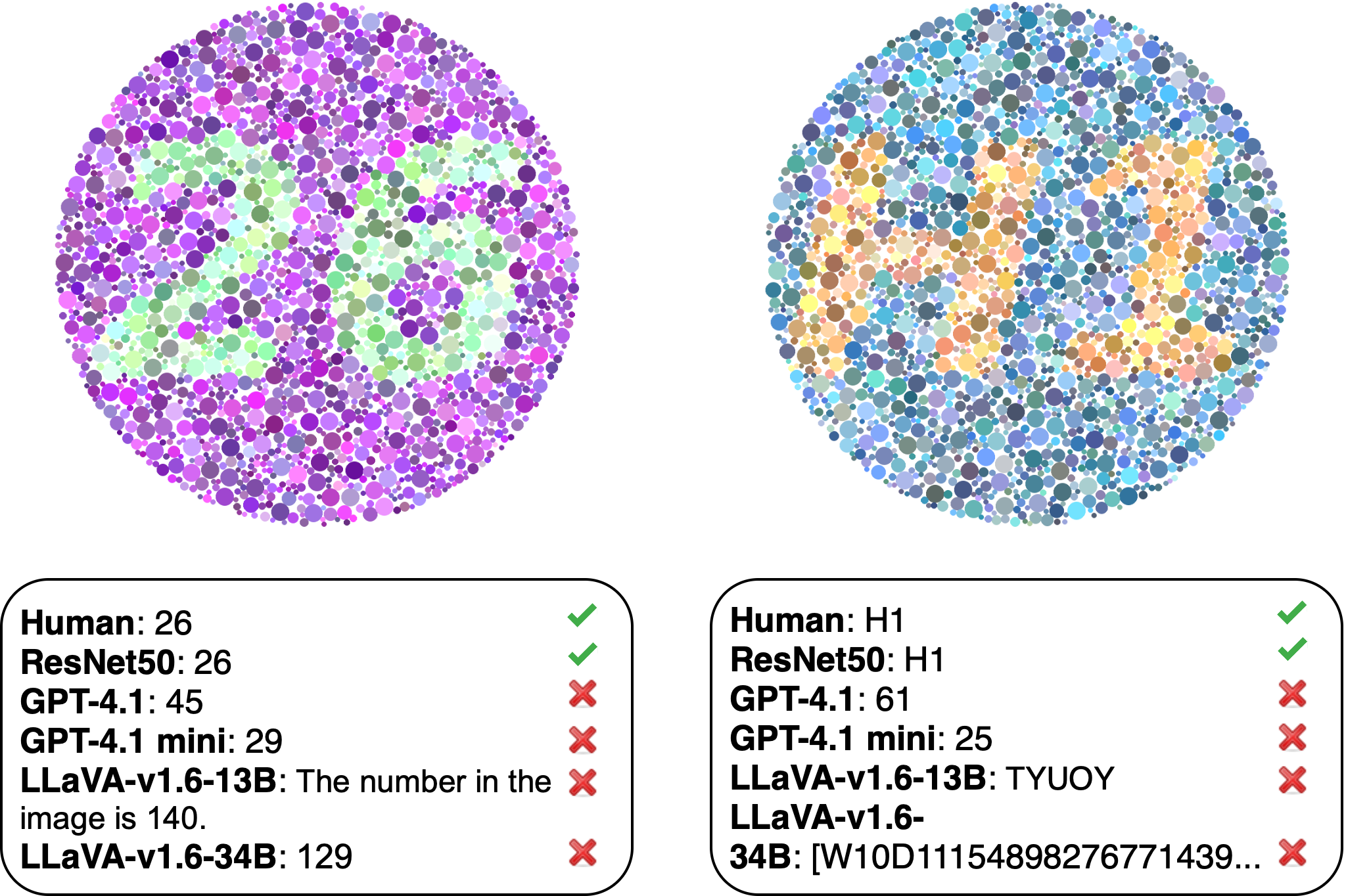}
    \caption{Qualitative examples showing predictions of 4 representative MLLMs vs baselines on numeric and alphanumeric tasks.}
    \label{fig:ishihara-pattern}
\end{figure}

\textbf{Ishihara-Style Pattern Generation.}
Our pattern generator, adapted from an open-source Pygame project\footnote{\url{https://github.com/hakrackete/Ishihara-color-plate-generator}}, iteratively populates the image with non-overlapping circles. Over 30,000 iterations, the generator randomly places circles, computes their maximum non-colliding radius (4-15 pixels), and assigns it a color. The color assignment depends on whether the circle's center falls within the character region of the text mask. This initial color then undergoes three randomized transformations: a gradient shift towards the other color, an RGB color shift (range [-30, +30]), and an RGB lightness scaling (factor 0.66 - 1.5). These transformed circles are then rendered, resulting in the final dense Ishihara-style pattern (Figure \ref{fig:ishihara-pattern}).

\section{Experimental Setup} \label{experiments}
\textbf{Benchmark Tasks and Evaluation Sets.}
From the generated stimuli, we designed a benchmark with two recognition tasks of varying difficulty and two visual conditions to isolate specific model capabilities. There are two distinct recognition tasks of varying difficulty: a numeric task with a label space of 90 possible outputs and a more challenging alphanumeric task with 3,364 possible outputs (Refer to Appendix B.1). The larger search space makes the alphanumeric task inherently more difficult.

\begin{itemize}[label=$\circ$]
    \item \textit{Number Recognition Set (Easier Task)}: This subset contains images with two-digit numeric strings (e.g., 17, 83, 65). We exclude leading-zero numbers (`00'-`09') to prevent ambiguity between single and double-digit interpretations.
    \item \textit{Alphanumeric Recognition Set (Harder Task)}: This subset includes two-character strings (e.g., A7, 9b, XG) formed from lowercase letters (a-z), uppercase letters (A-Z), and digits (0-9). We excluded visually ambiguous characters (`l', `I', `J', `O') from the character space.
\end{itemize}

Further, to disentangle the challenge of perceptual grouping from fundamental character recognition, we evaluate models under two visual conditions for each image. Due to computational and API cost constraints, our MLLM evaluations were conducted on subsets of 1,000 images, randomly sampled for both the numeric and alphanumeric tasks.
\begin{itemize}[label=$\circ$]
    \item \textit{Ishihara Pattern:} Primary task using the Ishihara-style dot pattern images.
    \item \textit{Text Masks:} Control task using the binary text mask used in the creation of Ishihara Pattern images. This condition establishes a baseline for each model's fundamental OCR capability, helping to disentangle it from performance on the more perceptually complex patterns.
\end{itemize}

\textbf{Multimodal Large Language Models (MLLMs).}
We evaluated a diverse set of nine Multimodal Large Language Models (MLLMs), including both commercial API-based and publicly available open-source models. Model inference was orchestrated using Promptfoo\footnote{\url{https://www.promptfoo.dev}}, a platform facilitating reproducible benchmarking through flexible prompt definition and API integration. Open-source models were hosted locally via Ollama\footnote{\url{https://ollama.com}} and inferred on a single NVIDIA A100 GPU. Images were Base64 encoded at 900x900 original resolution and submitted with the following task-specific prompts, which were kept consistent across all models and conditions:

% \begin{verbatim}
% What is the number in this image? Strictly stick to the format: Answer: {[}number in the image{]}
% \end{verbatim}

% \begin{itemize}
%     \item \textit{Number Recognition Prompt:} ``What is the number in this image? Strictly stick to the format: Answer: {[}number in the image{]}''
%     \item \textit{Text Recognition Prompt:} ``What is the exact text in this image? It has only alpha-numeric characters excluding small L, capital O, capital I, and capital J to avoid ambiguity. Strictly stick to the format: Answer: {[}exact text in the image{]}''
% \end{itemize}

\begin{tcolorbox}
    \textbf{Number Recognition Prompt:}
    \small\textit{``What is the number in this image? Strictly stick to the format: Answer: {[}number in the image{]}''}

    \vspace{2mm} % Adds a little space between the prompts

    \textbf{Text Recognition Prompt:}
    \small\textit{``What is the exact text in this image? It has only alpha-numeric characters excluding small l, capital O, capital I, and capital J to avoid ambiguity. Strictly stick to the format: Answer: {[}exact text in the image{]}''}
\end{tcolorbox}

\textbf{Human Performance Baseline.}
To establish a human performance baseline, we recruited 16 adult volunteers from diverse age groups with self-reported normal color vision. Each participant was evaluated on 100 Ishihara pattern images for each of the tasks (numerical and alphanumeric recognition). These images were randomly sampled from the 1,000-image sets used in the MLLM evaluations. Participants viewed the 900x900 pixel images in a Google Sheets document and provided their responses to the same prompts given to the models, ensuring a direct and methodologically consistent comparison.

\textbf{Traditional Computer Vision Baseline (ResNet50).} \label{sec:resnet_methodology}
As a representative traditional computer vision baseline, we trained and evaluated a ResNet50 model. We utilized a ResNet50 pre-trained on ImageNet, obtained from the PyTorch \texttt{vision} library. The standard classification layer of the ResNet50 was replaced with two independent classification heads. Each head was designed to predict one character, treating the task as two independent character recognition problems. For the purpose of fine-tuning this model, we utilized 2,000 images randomly sampled from the broader HueManity dataset, ensuring these were distinct from the final evaluation subsets. Training was conducted for 30 epochs using the Adam optimizer with a learning rate of $1e-3$. The loss function was the sum of the cross-entropy losses calculated independently for each of the two classification heads. The trained model was evaluated on the same 1,000-image subsets used for the MLLM evaluations.

\section{Results and Analysis} \label{sec:results_and_analysis}

\begin{table}[h]
  \caption{\textbf{Accuracy on the number and alphanumeric recognition tasks} for human evaluators, ResNet50, and various MLLMs on both text masks and patterned HueManity images. $\pm$ denotes the Wilson Confidence intervals at 95\% confidence.} 
  \label{model-num-eval-results}
  \centering
  \resizebox{\textwidth}{!}{%
  \begin{tabular}{lcccc}
    \toprule
    & \multicolumn{2}{c}{\textbf{{Number Task}}} & \multicolumn{2}{c}{\textbf{Alphanumeric Task}} \\
     & \textbf{Mask} & \textbf{Pattern} & \textbf{Mask} & \textbf{Pattern}\\
     \toprule
     Random choice & 1.11\% & 1.11\% & 0.029\% & 0.029\% \\
    \toprule
    Humans (average) & - & 99.37 $\pm$ 2.37\% & - & 93.25 $\pm$ 5.08\% \\
    ResNet50 & - & 96.5 $\pm$ 1.15\% & - & 94.5 $\pm$ 1.42\%\\
    \midrule
    \multicolumn{5}{c}{\small{API-based models}} \\
    \midrule
    GPT-4.1 & 100 $\pm$ 0.19\% & 33.6 $\pm$ 2.92\% & 80 $\pm$ 2.47\% & 3.0 $\pm$ 1.07\% \\
    GPT-4.1 mini & 100 $\pm$ 0.19\% & 19.0 $\pm$ 2.42\% & 72.4 $\pm$ 2.76\% & 0.6 $\pm$ 0.51\% \\
    Claude 3.7 Sonnet & 100 $\pm$ 0.19\% & 0.4 $\pm$ 0.43\% & 82.2 $\pm$ 2.36\% & 0 $\pm$ 0.19\%\\
    \midrule
    \multicolumn{5}{c}{\small{Open-source models}} \\
    \midrule
    LLaVA-v1.6-34B & 96.6 $\pm$ 1.13\% & 7.8 $\pm$ 1.66\% & 27.1 $\pm$ 2.75\% & 0 $\pm$ 0.19\%\\
    LLaVA-v1.6-13B & 87.2 $\pm$ 2.07\% & 8.1 $\pm$ 1.69\% & 31.8 $\pm$ 2.88\% & 0.1 $\pm$ 0.27\%\\
    LLaVA-v1.6-7B & 87.7 $\pm$ 2.03\% & 3.3 $\pm$ 1.11\% & 15 $\pm$ 2.21\% & 0 $\pm$ 0.19\%\\
    Mistral-small3.1-24b & 100 $\pm$ 0.19\% & 0.1 $\pm$ 0.27\% & 58.7 $\pm$ 3.04\% & 0 $\pm$ 0.19\%\\
    Qwen VL Max & 100 $\pm$ 0.19\% & 0.2 $\pm$ 0.33\% & 83.5 $\pm$ 2.29\% & 0 $\pm$ 0.19\%\\
    Pixtral & 100 $\pm$ 0.19\% & 1 $\pm$ 0.64\% & 65.8 $\pm$ 2.93\% & 1.8 $\pm$ 0.84\% \\
    \bottomrule
  \end{tabular}}
\end{table}

\subsection{Human and ResNet50 Baseline Performance} 
\label{sec:baselines_eval_results} 

To contextualize the performance of MLLMs, we established two critical baselines: human evaluators and a fine-tuned ResNet50 model. These baselines serve to confirm the task's solvability and provide a clear performance ceiling.

Human evaluators demonstrated near-perfect accuracy, achieving 99.37\% on the numerical task and 93.25\% on the alphanumeric task. Annotator feedback indicated that the few errors on the alphanumeric task were due to confusion between visually similar characters (e.g., `s' vs. `S', `c' vs. `C', `w' vs. `W'), not an inability to perceive the patterns. Annotators also report that the character recognition time was in the order of a few seconds. This near-flawless performance confirms that the stimuli are clear and the task is fundamentally solvable for a proficient visual system.

Similarly, a fine-tuned ResNet50 \cite{he2015deepresiduallearningimage} model achieved high accuracy, scoring 96.5\% on the numerical task and 94.5\% on the alphanumeric task. This strong performance from a standard convolutional architecture, trained on only 2,000 examples, proves that the perceptual cues within the images are sufficient for established computer vision techniques to learn the task.

Together, these baselines establish that the HueManity benchmark is a well-posed and learnable perception challenge, suggesting that any failures by MLLMs are likely due to architectural or training-related limitations rather than the inherent difficulty of the task itself.

\subsection{MLLM Performance}

The performance of the nine evaluated MLLMs on HueManity sharply contrasts with the near-perfect accuracies of both humans and the ResNet50 baseline (Table \ref{model-num-eval-results}). All MLLMs consistently struggled, with the best achieving only 33.6\% on the numeric task and a mere 3\% on the alphanumeric task. 

As expected, most models demonstrated strong Optical Character Recognition (OCR) capabilities on the simple binary mask images. For the numeric mask task, nearly all models achieved perfect or near-perfect accuracy. Performance on the more complex alphanumeric mask was more varied, while API-based models like GPT-4.1 and Claude 3.7 Sonnet maintained high accuracy ($>$80\%), the open-source LLaVA family notably struggled (15-32\% accuracy), indicating a weaker baseline OCR capability for complex character sets.

The models' high OCR accuracy on masks makes their widespread failure on the patterned images even more striking. This drop in performance pinpoints the challenge as a perceptual failure, not a simple recognition one. On the ``easy'' numeric pattern task, a clear performance hierarchy emerged. GPT-4.1 was the only model with meaningful success at 33.6\% accuracy. A second tier, including GPT-4.1 mini (19.0\%) and the LLaVA family (3.3-8.1\%), showed minimal capability. The remaining models, such as Claude 3.7 Sonnet and Qwen VL Max, failed almost completely, scoring near zero. On the ``hard'' alphanumeric pattern task, MLLM performance collapsed almost entirely. GPT-4.1 was again the top performer but with a mere 3.0\% accuracy. It was followed by Pixtral at 1.8\%. All other evaluated models scored less than 1\%, demonstrating a near-total inability to perform perceptual grouping on this more complex task.

Several characteristics inherent to the current design and training paradigms of many MLLMs may contribute to their observed difficulties on tasks demanding nuanced visual perception.

\textbf{Semantic Optimization in Pre-Training.} MLLM vision encoders are typically optimized to capture high-level semantic information for tasks like scene understanding \cite{llava, liu2024improved, agrawal2024pixtral12b, qwenvl, bai2025qwen25vltechnicalreport}. This focus on global context may cause the loss of fine-grained local details, such as the subtle color and texture cues that define the characters in HueManity. Furthermore, pre-training on web-scale datasets of images paired with descriptive text may not provide sufficient exposure to stimuli requiring intensive perceptual organization without strong semantic anchors.

\textbf{Architectural Bottlenecks.} The architectures themselves may contribute to this failure. Many MLLMs use Vision Transformers (ViTs) \cite{dosovitskiy2021imageworth16x16words} that divide images into patches, a process that can disrupt fine-grained details within those patches. Additionally, the projection layers that connect the vision encoder to the language model can act as an information bottleneck, abstracting away precise, high-resolution feature distinctions that are essential for solving our task.

\subsection{Can MLLMs Learn the Task? Probing with In-Context Learning and Fine-Tuning}

\textbf{In-Context Learning}
We tested whether providing in-context examples could improve performance. While this approach retained 100\% accuracy for the simple ``mask'' images, it proved ineffective and often detrimental for the patterned Ishihara plates (Table \ref{few-shot-learning-results}). For most models, performance degraded as more examples were added. These results are consistent with \cite{illusionvqa}'s findings and we hypothesize that the visual complexity of the examples introduces noise that confuses the models, reinforcing the conclusion that the deficit is perceptual rather than a lack of contextual understanding.

\begin{table}[h!]
  \caption{Few-shot performance on Ishihara Pattern for number task with 95\% Wilson confidence intervals.}
  \label{few-shot-learning-results}
  \centering
  \resizebox{0.9\textwidth}{!}{%
  \begin{tabular}{lcccc}
    \toprule
     \textbf{Model} & \textbf{0-shot (Baseline)} & \textbf{1-shot} & \textbf{3-shot} & \textbf{5-shot} \\
    \toprule
    GPT 4.1 & 33.6 $\pm$ 2.9\% & 34.0 $\pm$ 2.9\% & 35.0 $\pm$ 3.0\% & 31.0 $\pm$ 2.9\% \\
    GPT-4.1 mini & 19.0 $\pm$ 2.5\% & 14.0 $\pm$ 2.2\% & 9.0 $\pm$ 1.8\% & 6.0 $\pm$ 1.5\% \\
    Pixtral Large & 1.0 $\pm$ 0.6\% & 1.0 $\pm$ 0.6\% & 0.0 $\pm$ 0.2\% & 0.0 $\pm$ 0.2\% \\
    Claude 3.7 Sonnet & 0.4 $\pm$ 0.4\% & 0.0 $\pm$ 0.2\% & 1.0 $\pm$ 0.6\% & 1.0 $\pm$ 0.6\% \\
    Qwen 2.5 VL Max & 0.2 $\pm$ 0.3\% & 0.0 $\pm$ 0.2\% & 0.0 $\pm$ 0.2\% & 0.0 $\pm$ 0.2\% \\
    \bottomrule
  \end{tabular}}
\end{table}

\begin{table}[h!]
  \caption{Performance of fine-tuned Gemma3-4B and qualitative examples of predictions.}
  \label{mllm-finetuning-results}
  \centering
  \begin{tabularx}{1.0\textwidth}{@{} llr *{4}{>{\centering\arraybackslash}X} @{}}
    \toprule
    % Main header row
    \multirow{2}{*}{\textbf{Training Data}} & 
    \multirow{2}{*}{\textbf{Task}} & \multirow{2}{*}{\textbf{Accuracy}} & 
    \multicolumn{4}{c}{\textbf{Predictions for Ground Truth Values}} \\
    \cmidrule(l){4-7}
    & & & \small ML & \small sZ & \small cT & \small 1Q \\
    \midrule
    
    % --- Zero-Shot ---
    \multirow{2}{*}{N/A (Zero-Shot)} 
    & Mask & 80\% & \small ML & \small SZ & \small CT & \small 1Q \\
    & Pattern & 0\% & \small 76492385 & \small 739482561 & \small 74928536 & \small 749238561 \\
    \midrule
    
    % --- Fine-Tuned on Masks ---
    \multirow{2}{*}{Fine-Tuned on Masks} 
    & Mask & 94\% & \small ML & \small sZ & \small cT & \small 1Q \\
    & Pattern & 0\% & \small R1 & \small M5 & \small 4model & \small D4 \\
    \midrule
    
    % --- Fine-Tuned on Ishihara ---
    \multirow{2}{*}{Fine-Tuned on Ishihara} 
    & Mask & 0\% & \small 7s & \small 24 & \small n6 & \small 36 \\
    & Pattern & 0\% & \small tH & \small vF & \small 6t & \small 58\\
    
    \bottomrule
  \end{tabularx}
\end{table}

\textbf{MLLM Fine-tuning}
To determine if direct training could overcome the observed perceptual challenges, we LoRA fine-tuned the Gemma-3-4B model on 500 examples from the HueManity dataset using the HuggingFace SFTTrainer\footnote{\url{https://ai.google.dev/gemma/docs/core/huggingface_text_full_finetune}} with Adam optimizer, 2e-4 learning rate, and 3 epochs. The results of this limited experiment were striking and revealed a critical limitation. While fine-tuning on the simple text masks improved performance on that control task as expected, it provided no benefit for recognizing the patterned Ishihara images. More critically, fine-tuning on the patterned images not only failed to improve performance on the core task but also destroyed the model's ability to solve the simple mask task, causing its accuracy to plummet from 80\% to 0\%. Instead of learning to perceive the characters, the model simply learned to output plausible-looking two-character strings (e.g., ``4G'', ``6t'') regardless of the visual input. This suggests the failure is rooted in deeper architectural limitations, as the model appears incapable of learning the necessary perceptual skill.

% \begin{table}[h]
%   \caption{Fine-tuning Experiment on Gemma3 4B}
%   \label{mllm-finetuning-results}
%   \centering
%   \begin{tabularx}{0.9\textwidth}{@{} l cc X @{}} 
%     \toprule
%     \textbf{Training Data} & \multicolumn{2}{c}{\textbf{Alphanumeric Task}} & \textbf{Observation}\\
%     & \textbf{Mask} & \textbf{Pattern} & \\
%     \midrule
%     N/A (Zero-Shot) & 80.00\% & 0.00\% & The base model reads the masks but fails on the Ishihara plates. \\
%     Fine-Tuned on Masks & 94.00\% & 0.00\% & Performance on the masks improves as expected. Fine-tuning on masks provides no benefit for the Ishihara task. \\
%     Fine-Tuned on Ishihara & 0.00\% & 0.00\% & The model unlearns how to read the masks and still fails to perceive the patterned plates. \\
%     \bottomrule
%   \end{tabularx}
% \end{table}

\subsection{Ablations and Additional Analyses}

\textbf{Image Resolution Ablation.} To understand the impact of image resolution, we conducted an ablation study by evaluating model performance across a range of resolutions from 300px to 1300px (Table \ref{image-resolution-results}). The results were twofold. First, models that performed poorly at the native 900px resolution, such as Qwen VL Max, failed across all tested resolutions, which reinforces the claim that they have a fundamental perceptual deficit. Second, we uncovered a ``squinting effect'' in the best-performing model, GPT-4.1, whose accuracy peaked at the lowest resolution, increasing from 34\% to 49\% at 300px. We hypothesize that downsampling acts as a low-pass filter, smoothing the high-frequency ``noise'' from the dot patterns and making the underlying character shapes more salient to the model.

\begin{table}[h]
  \caption{Image resolution ablation with 95\% Wilson confidence intervals (N=100).}
  \label{image-resolution-results}
  \centering
  \resizebox{\textwidth}{!}{%
  \begin{tabular}{lcccccc}
    \toprule
     \textbf{Model} & \textbf{300px} & \textbf{500px} & \textbf{700px} & \textbf{900px} & \textbf{1100px} & \textbf{1300px}\\
    \toprule
    GPT 4.1 & $49.0 \pm 9.8\%$ & $29.0 \pm 8.9\%$ & $31.0 \pm 9.1\%$ & $34.0 \pm 9.3\%$ & $30.0 \pm 9.0\%$ & $29.2 \pm 8.9\%$ \\
    GPT-4.1 mini & $39.0 \pm 9.6\%$ & $18.2 \pm 7.6\%$ & $18.0 \pm 7.6\%$ & $18.0 \pm 7.6\%$ & $10.0 \pm 5.9\%$ & $6.0 \pm 4.6\%$ \\
    Qwen VL Max & $1.0 \pm 2.0\%$ & $1.0 \pm 2.0\%$ & $1.0 \pm 2.0\%$ & $1.0 \pm 2.0\%$ & $1.0 \pm 2.0\%$ & $1.0 \pm 2.0\%$ \\
    Pixtral Large & $1.0 \pm 2.0\%$ & $0.0 \pm 1.9\%$ & $1.0 \pm 2.0\%$ & $0.0 \pm 1.9\%$ & $0.0 \pm 1.9\%$ & $0.0 \pm 1.9\%$ \\
    \bottomrule
  \end{tabular}}
\end{table}

\textbf{MLLM Failure Patterns.} A qualitative analysis of MLLM failures on the HueManity benchmark reveals that, unlike humans whose errors are predictable, MLLMs exhibit three distinct and more fundamental failure patterns when their perceptual abilities are overwhelmed (see Appendix E for more details).
\begin{itemize}[label=$\circ$]
    \item Hallucination: Models often generate completely unrelated and overly complex text, such as turning a two-character string into a full phrase (e.g., `MUST SEE'). This happens when the perceptual challenge overwhelms their visual processing.
    \item Evasion: They frequently engage in evasive behavior, either by describing the image as an `Ishihara test' without attempting to identify the characters or by explicitly stating they are unable to perform the task. This suggests an internal confidence threshold triggers a pre-programmed failure response.
    \item Erratic Outputs: Their outputs are often erratic and unpredictable, ranging from random strings and nonsensical numbers to flawed, repetitive patterns (e.g., `[L1L1L1]'). This unpredictability points to a lack of robust and stable visual feature extraction.
\end{itemize}

\textbf{Quantitative Correlation with Real-World Benchmarks.}
To validate that performance on HueManity translates to real-world capabilities, we correlated model rankings on our alphanumeric pattern task with the human-preference Elo ratings from Vision Arena \cite{chou2025visionarena230krealworld}, a benchmark for diverse, real-world multimodal tasks. Our analysis included the seven models common to both leaderboards: GPT-4.1, GPT-4.1 mini, Claude 3.7 Sonnet, Mistral Medium, Qwen VL Max, Pixtral Large, and LLaVA-v1.6-34B. The analysis revealed a \textbf{strong and statistically significant positive correlation} of $\bm{\rho = 0.8214}$ ($\bm{p=0.0227}$). This result provides compelling empirical evidence that HueManity is not an isolated synthetic challenge but effectively measures a foundational perceptual skill that is highly predictive of an MLLM's general performance on complex, real-world tasks. This validates our benchmark as a potent diagnostic tool for identifying critical and generalizable gaps in the core perceptual abilities of current models.

% \subsubsection{Qualitative Analysis in Cluttered Natural Images}
% To qualitatively assess this link, we created a small diagnostic set of 100 real-world images where text is partially obscured by visual noise (e.g., rain, glare, textures). Preliminary tests show that models struggling with HueManity also fail to reliably perform optical character recognition (OCR) in these cluttered natural scenes, whereas models with higher HueManity scores perform better. This provides further qualitative evidence that our benchmark effectively isolates a core perceptual skill essential for robust real-world performance.

\section{Conclusion and Future Directions} \label{conclusions}
In this work, we introduce HueManity, a large-scale benchmark with 83,850 Ishihara-style images designed to probe the fine-grained visual perception of Multimodal Large Language Models (MLLMs). Our comprehensive evaluation of nine state-of-the-art models reveals a critical weakness: the best-performing MLLM achieved only (33.6\%, 3\%) accuracies on the numeric and alphanumeric Ishihara pattern recognition tasks, starkly contrasting with the near-perfect scores of human evaluators (99.37\%, 93.25\%) and a fine-tuned ResNet50 baseline (96.5\%, 94.5\%). We demonstrate that this perceptual deficit is not easily remedied, as neither in-context learning nor direct fine-tuning yielded improvements, suggesting the failure may be rooted in fundamental architectural limitations rather than the task's inherent difficulty. The strong correlation ($\rho = 0.82$) between HueManity performance and rankings on the real-world Vision Arena benchmark validates that our findings are not isolated to a synthetic task but are indicative of a broader, more generalizable weakness in MLLM perceptual abilities. 

The deficiencies observed likely stem from vision encoders optimized for high-level semantics at the expense of fine-grained detail, information bottlenecks at the vision-language interface, and pre-training datasets that lack sufficient perceptual challenges. To bridge this gap, future work should focus on developing novel MLLM architectures that preserve low-level visual information, augmenting training datasets with perceptually challenging stimuli, and designing training objectives that explicitly foster robust visual acuity independent of high-level semantic reasoning. We release our dataset and generation code to facilitate further research toward building more perceptually grounded MLLMs.

% \input{ICPR_2026_LaTeX_Templates/sections/limitations}

%
% ---- Bibliography ----
%
% BibTeX users should specify bibliography style 'splncs04'.
% References will then be sorted and formatted in the correct style.
%
\bibliographystyle{splncs04}
\bibliography{ICPR_2026_LaTeX_Templates/huemanity}

\appendix
\section{Model Inventory} \label{sec:model_inventory}

\begin{table*}[h]
\centering
\small
\resizebox{0.95\textwidth}{!}{%
\begin{tabular}{l l l l l}
\toprule
\textbf{Model} & \textbf{Provider} & \textbf{Version} & \textbf{Temp / Top-p} & \textbf{Max Tokens} \\
\midrule
% ---- FILLED ROWS ----
GPT-4.1 mini & OpenAI & gpt-4.1-mini-2025-04-14 & 1.0 / 1.0 & 1024 \\
GPT-4.1 & OpenAI & gpt-4.1-2025-04-14 & 1.0 / 1.0 & 1024 \\
Claude 3.7 Sonnet & Anthropic & claude-3-7-sonnet-20250219 & 1.0 / 1.0 & 1024 \\
LLaVA-v1.6-7B & Ollama & llava:7b & 0.8 / 0.9 & -1 (inf) \\
LLaVA-v1.6-13B & Ollama & llava:13b & 0.8 / 0.9 & -1 (inf) \\
LLaVA-v1.6-34B & Ollama & llava:34b & 0.8 / 0.9 & -1 (inf) \\
Mistral-small3.1-24b & Mistral AI & mistral-small-3.1-24b-instruct & 0.7 / 1.0 & 1024 \\
Qwen VL Max & Qwen (OpenRouter) & qwen/qwen-vl-max & 1.0 / 1.0 & 1024 \\
Pixtral & Mistral AI & pixtral-large-2411 & 0.7 / 1.0 & 1024 \\
\bottomrule
\end{tabular}}
\caption{Standardized evaluation settings per model. All models were evaluated between May 6-7, 2025. Ollama models were hosted on a single GPU machine with an NVIDIA A100. These generation values are the model/provider defaults and we stick to these values to avoid introducing noise to our evaluations.}
\label{tab:model_settings}
\end{table*}

\section{Label Spaces \& Chance Baselines}
\label{app:label-space}

\paragraph{Character universe.}
We define a character set \( \mathcal{C} \) used to compose two-character labels.
Starting from decimal digits and ASCII letters with case,
\[
\{0,\ldots,9,\,A,\ldots,Z,\,a,\ldots,z\}\quad (62\ \text{symbols}),
\]
we exclude four visually ambiguous characters:
lowercase \(\ell\) and uppercase \(\{I, J, O\}\).
Thus,
\[
|\mathcal{C}| = 62 - 4 = 58.
\]

\subsection{Tasks and exact label spaces} \label{sec:label-space}

\paragraph{Numeric (two digits).}
The numeric label space contains all two-digit numbers \emph{without a leading zero}:
\[
\mathcal{N}=\{10,11,\ldots,99\}, \qquad |\mathcal{N}|=90.
\]
A uniform random-guess baseline is therefore
\[
\Pr[\text{random correct}] = \frac{1}{|\mathcal{N}|} = \frac{1}{90} \approx 1.11\%.
\]

\paragraph{Alphanumeric (two characters).}
The alphanumeric label space contains \emph{all ordered pairs} from \(\mathcal{C}\):
\[
\mathcal{A}=\mathcal{C}\times \mathcal{C}, \qquad |\mathcal{A}| = 58^2 = 3364.
\]
A uniform random-guess baseline is therefore
\[
\Pr[\text{random correct}] = \frac{1}{|\mathcal{A}|} = \frac{1}{3364} \approx 0.0297\%.
\]

\subsection{Scoring policy}
We compute case-sensitive exact-match accuracy over \(\mathcal{N}\) or \(\mathcal{A}\), as applicable.
Normalization prior to scoring trims whitespace, removes a possible ``Answer: " prefix, extracts the first two alphanumerics.
Characters outside \(\mathcal{C}\) are considered invalid.

\section{A Brief Discussion on CIEDE2000 Color Difference} \label{sec:CIEDE2000}
\begin{figure*}[h]
\begin{align} \label{deltae2000_equation}
\Delta E_{2000} &= \sqrt{
    \left( \frac{\Delta L'}{K_L S_L} \right)^2 +
    \left( \frac{\Delta C'}{K_C S_C} \right)^2 +
    \left( \frac{\Delta H'}{K_H S_H} \right)^2 +
    R_T \left( \frac{\Delta C'}{K_C S_C} \right)
    \left( \frac{\Delta H'}{K_H S_H} \right)
} \nonumber \\
&\begin{aligned}
\text{where} \quad
&\Delta L' \text{ is the corrected lightness difference,} \\
&\Delta C' \text{ is the corrected chroma difference,} \\
&\Delta H' \text{ is the corrected hue difference,} \\
&K_L, K_C, K_H \text{ are parametric factors (typically 1),} \\
&S_L, S_C, S_H \text{ are weighting functions for lightness, chroma, and hue,} \\
&R_T \text{ is a rotation term accounting for hue-chroma interaction.}
\end{aligned}
\end{align}
\end{figure*}

\begin{figure*}
    \centering
    \includegraphics[width=0.8\linewidth]{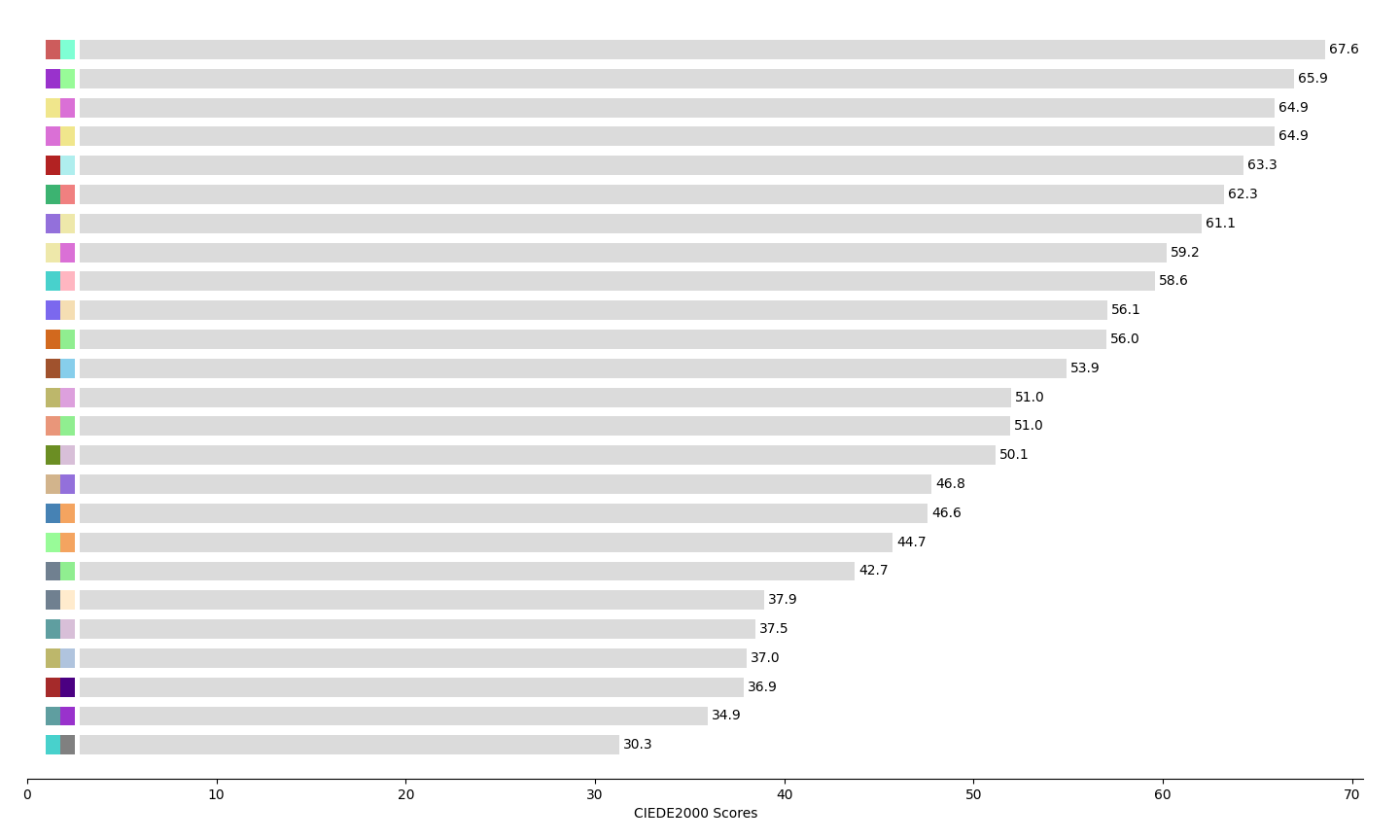}
    \caption{Distribution of CIEDE2000 color difference scores for the 25 selected foreground-background color pairs utilized in the HueManity benchmark.}
    \label{fig:score-distribution}
\end{figure*}

The CIEDE2000 score ($\Delta E_{2000}$, Equation~\ref{deltae2000_equation}) \cite{article} quantifies the perceived difference between two colors more accurately than prior formulae, especially for subtle variations. It calculates a single value representing the ``distance'' between colors in the perceptually uniform CIE $L^*a^*b^*$ space, considering lightness, chroma, and hue. In the HueManity benchmark, $\Delta E_{2000}$ was pivotal for systematically designing stimuli. The ability to discern characters in the Ishihara-style plates directly depends on the perceived color contrast between foreground (character) and background dots. This score provided a perceptually relevant, objective method to quantify this contrast, enabling the selection of color pairs across a controlled spectrum of difficulty, refer to Figure~\ref{fig:score-distribution}. This ensured the benchmark could rigorously test visual perception for varying degrees of color discriminability while maintaining stimuli legibility for human comparison, forming a foundational aspect of our dataset's controlled experimental design.

\section{Color Pairs Selection} \label{sec:color selection}
The selection of appropriate color pairs for the foreground (characters) and background dots was a critical phase in the development of HueManity, undertaken with considerable care to ensure a balance between perceptual challenge and unambiguous human legibility. The process involved several stages:
\begin{enumerate}
\item \textbf{Initial Candidate Generation:} We bootstrapped the process with ~15 medium-contrast color pairs generated by LLMs (Gemini, ChatGPT). This initial pool was iteratively refined by evaluating pairs against CIEDE2000 ($\Delta E_{2000}$, Eq. \ref{deltae2000_equation}) scores and visual checks. We retained promising candidates, modified some, and discarded others, while simultaneously manually crafting and vetting new pairs to meet the benchmark's final requirements (detailed below). This refinement cycle culminated in the selection of 25 distinct pairs for the subsequent validation stages.
\item \textbf{Quantitative Contrast Filtering (CIEDE 2000):} Each of these candidate pairs then underwent rigorous quantitative analysis using the CIEDE2000 ($\Delta E_{2000}$) color difference formula (Equation \ref{deltae2000_equation}). This formula is a standard measure in color science, designed to reflect perceptually meaningful differences as perceived by humans. We established a specific target range for the $\Delta E_{2000}$ score, retaining only pairs with contrast values between 25 and 75. The lower bound of 25 was set to ensure sufficient theoretical distinguishability for individuals with normal color vision, preventing pairs that would be inherently ambiguous. The upper bound of 75 aimed to exclude pairs with excessively high contrast, which might render the perceptual task trivial and deviate from the subtle challenge intended.
\item \textbf{Balanced Contrast Distribution:} A key objective during selection was to ensure the benchmark included stimuli across a spectrum of difficulty levels related to color similarity. Therefore, we deliberately curated the final set of 25 color pairs to achieve an approximately equal distribution around a $\Delta E_{2000}$ score of 50. This threshold is grounded in color science literature, often considered a point distinguishing more subtle (scores $<$ 50) from more clearly distinct (scores $>$ 50) color differences. We aimed for roughly half the selected pairs to fall below this threshold and half above, ensuring HueManity evaluates performance across varying, literature-informed degrees of color contrast difficulty.
\item \textbf{Manual Verification and Legibility Check:} Recognizing that a single numerical contrast score like $\Delta E_{2000}$ captures overall perceived difference but may not fully account for the complex interplay of hue, saturation, and luminance components, especially when rendered as dots and subjected to further transformations (gradient, color, and light shifts), a crucial final step of manual verification was performed. It is hard to quantify the nuanced visual impact of these combined factors with a single metric, therefore for every color pair that passed the quantitative filtering, sample HueManity images were generated. These renderings were meticulously inspected by the authors. The primary goal was to reject pairs where the characters, despite an acceptable overall contrast score, appeared visually too similar to the background due to the specific combination of hue, saturation, luminance, or the effect of the applied shifts. This ensured that the embedded alphanumeric characters were clearly legible and that the pattern recognition was unambiguous for human observers with normal color vision. Any pairs that resulted in ambiguous characters or were otherwise problematic during this visual check were discarded.
\end{enumerate}

This multi-stage process, combining LLM-based idea generation, principled quantitative filtering based on color science, a balanced distributional strategy, and crucial human judgment to account for complex visual interactions, resulted in the final curated set of 25 color pairs. This ensures that the stimuli used in HueManity are not only theoretically sound but also practically validated for fairness, legibility, and the intended level of perceptual challenge.

\begin{table*}[h!]
\centering
\caption{The 25 Curated Color Pairs Used in the HueManity Benchmark.}
\label{tab:color-pairs}
\resizebox{1.0\textwidth}{!}{%
\begin{tabular}{l l l}
\toprule
\textbf{Foreground Color} & \textbf{Background Color} & \textbf{(Foreground | Background) RGB Values} \\
\midrule
\cellcolor{cFirebrick}{\textcolor{white}{Firebrick}} & \cellcolor{cPaleTurquoise}{PaleTurquoise} & (178, 34, 34) | (175, 238, 238) \\
\cellcolor{cSienna}{\textcolor{white}{Sienna}} & \cellcolor{cSkyBlue}{SkyBlue} & (160, 82, 45) | (135, 206, 235) \\
\cellcolor{cOliveDrab}{\textcolor{white}{OliveDrab}} & \cellcolor{cThistle}{Thistle} & (107, 142, 35) | (216, 191, 216) \\
\cellcolor{cDarkOrchid}{\textcolor{white}{DarkOrchid}} & \cellcolor{cPaleGreen}{PaleGreen} & (153, 50, 204) | (152, 251, 152) \\
\cellcolor{cSteelBlue}{\textcolor{white}{SteelBlue}} & \cellcolor{cSandyBrown}{SandyBrown} & (70, 130, 180) | (244, 164, 96) \\
\cellcolor{cMediumSeaGreen}{\textcolor{white}{MediumSeaGreen}} & \cellcolor{cLightCoral}{LightCoral} & (60, 179, 113) | (240, 128, 128) \\
\cellcolor{cChocolate}{Chocolate} & \cellcolor{cLightGreen}{LightGreen} & (210, 105, 30) | (144, 238, 144) \\
\cellcolor{cKhaki}{Khaki} & \cellcolor{cOrchid}{Orchid} & (240, 230, 140) | (218, 112, 214) \\
\cellcolor{cDarkKhaki}{DarkKhaki} & \cellcolor{cPlum}{Plum} & (189, 183, 107) | (221, 160, 221) \\
\cellcolor{cOrchid}{Orchid} & \cellcolor{cKhaki}{Khaki} & (218, 112, 214) | (240, 230, 140) \\
\cellcolor{cIndianRed}{\textcolor{white}{IndianRed}} & \cellcolor{cAquamarine}{Aquamarine} & (205, 92, 92) | (127, 255, 212) \\
\cellcolor{cPaleGreen}{PaleGreen} & \cellcolor{cSandyBrown}{SandyBrown} & (152, 251, 152) | (244, 164, 96) \\
\cellcolor{cBrown}{\textcolor{white}{Brown}} & \cellcolor{cIndigo}{\textcolor{white}{Indigo}} & (165, 42, 42) | (75, 0, 130) \\
\cellcolor{cCadetBlue}{\textcolor{white}{CadetBlue}} & \cellcolor{cDarkOrchid}{\textcolor{white}{DarkOrchid}} & (95, 158, 160) | (153, 50, 204) \\
\cellcolor{cPaleGoldenrod}{PaleGoldenrod} & \cellcolor{cOrchid}{Orchid} & (238, 232, 170) | (218, 112, 214) \\
\cellcolor{cMediumTurquoise}{MediumTurquoise} & \cellcolor{cGrey}{\textcolor{white}{Grey}} & (72, 209, 204) | (128, 128, 128) \\
\cellcolor{cSlateGrey}{\textcolor{white}{SlateGrey}} & \cellcolor{cLightGreen}{LightGreen} & (112, 128, 144) | (144, 238, 144) \\
\cellcolor{cMediumSlateBlue}{\textcolor{white}{MediumSlateBlue}} & \cellcolor{cWheat}{Wheat} & (123, 104, 238) | (245, 222, 179) \\
\cellcolor{cMediumTurquoise}{MediumTurquoise} & \cellcolor{cLightPink}{LightPink} & (72, 209, 204) | (255, 182, 193) \\
\cellcolor{cDarkSalmon}{DarkSalmon} & \cellcolor{cLightGreen}{LightGreen} & (233, 150, 122) | (144, 238, 144) \\
\cellcolor{cTan}{Tan} & \cellcolor{cMediumPurple}{\textcolor{white}{MediumPurple}} & (210, 180, 140) | (147, 112, 219) \\
\cellcolor{cMediumPurple}{\textcolor{white}{MediumPurple}} & \cellcolor{cPaleGoldenrod}{PaleGoldenrod} & (147, 112, 219) | (238, 232, 170) \\
\cellcolor{cSlateGray}{\textcolor{white}{SlateGray}} & \cellcolor{cBlanchedAlmond}{BlanchedAlmond} & (112, 128, 144) | (255, 235, 205) \\
\cellcolor{cCadetBlue}{\textcolor{white}{CadetBlue}} & \cellcolor{cThistle}{Thistle} & (95, 158, 160) | (216, 191, 216) \\
\cellcolor{cDarkKhaki}{DarkKhaki} & \cellcolor{cLightSteelBlue}{LightSteelBlue} & (189, 183, 107) | (176, 196, 222) \\
\bottomrule
\end{tabular}
}
\end{table*}

\section{Qualitative Analysis of MLLM Failure Patterns} \label{sec:qual_analysis}

This section details common failure patterns observed in Multimodal Large Language Models (MLLMs) when tasked with identifying alphanumeric characters embedded in Ishihara-style dot patterns from the HueManity dataset. These observations stem from a comparative analysis of MLLM responses against human performance and ground truth data. Notably, human visual perception proved highly accurate on these tasks, with any infrequent errors typically involving confusion between graphically similar characters. In contrast, MLLMs exhibited distinct and more fundamental failure modes.

\subsection{Prevalent Hallucination of Unrelated or Overly Complex Characters}

A dominant failure mode across multiple MLLMs was the generation of characters, words, or even entire phrases that bore no resemblance to the two-character ground truth. This phenomenon of ``hallucination'' often resulted in outputs significantly more complex or contextually incongruous than the target stimuli. For instance, in the alphanumeric task, a model such as \texttt{Claude 3.7 Sonnet} might interpret a simple two-letter combination as a short phrase (e.g., responding with ``MUST SEE'' or ``SOLU'' for simple targets like ``Rw'' or ``Tv''). Similarly, \texttt{llava-7b} could produce non-sensical strings like ``HQJHSTOS'', and \texttt{LLaVA-13b} occasionally generated contextually unrelated phrases like ``[G3T1NGST4RT3D]''. The numeric task was not immune --- for a two-digit number, \texttt{Claude 3.7 Sonnet} was observed to list a sequence of unrelated two-digit numbers. This pattern suggests that when the fine-grained perceptual challenge overwhelms the MLLMs' visual processing, they may default to generating text that, while perhaps linguistically plausible, is detached from the actual visual content.

\subsection{Frequent Resort to Descriptive Evasion or Explicit Admission of Inability}

Rather than consistently attempting to identify the embedded characters, many MLLMs frequently defaulted to one of two evasive strategies: providing a general description of the image (often correctly identifying it as a color vision test) or explicitly stating their incapacity to discern any characters. This behavior contrasted significantly with human participants, who invariably attempted the identification task. For example, models like \texttt{GPT-4.1 Mini} and \texttt{Mistral - small3.1 - 24b}, when presented with alphanumeric stimuli, often responded by describing the image as an Ishihara test but stated they could not clearly identify specific characters. In the numeric task, \texttt{Claude 3.7 Sonnet} sometimes offered similar descriptive evasions, asserting no number was visible and describing the circular dot pattern. Furthermore, some models, such as \texttt{LLaVA-34b}, occasionally provided categorical statements of inability, indicating they could not recognize or interpret images and requesting a description or textual input instead. This pattern suggests that MLLMs may possess internal confidence thresholds that, when triggered by low-confidence visual parsing, lead to evasive or pre-programmed ``unable to process'' responses rather than a forced, best-guess attempt at character recognition.

\subsection{Erratic, Unpredictable, and Systematically Flawed Output Patterns}

MLLM outputs were frequently characterized by their erratic and unpredictable nature. This included the generation of seemingly random strings of characters, peculiar systematic but incorrect patterns, or extreme numerical inventions far removed from the two-character target. This high variance in error types was observed both across different models for the same input and within the outputs of a single model across different images. For instance, when presented with the same alphanumeric target (e.g., ``Wh''), while one model (\texttt{GPT-4.1}) might respond almost correctly, others exhibited diverse failures: \texttt{Claude 3.7 Sonnet} produced an unrelated number (``4726''), \texttt{LLaVA-13b} generated an exceptionally long string of sequential numbers, and \texttt{Qwen VL Max} incorrectly reasoned the presence of a different number (``12''). Some incorrect outputs also suggested flawed systematic processing, such as \texttt{LLaVA-13b} responding with a patterned string like ``[L1L1L1]'' for one target or generating extremely long, patterned numeric strings for others. Lengthy, seemingly gibberish character strings were also common from models like \texttt{LLaVA-7b}. This unpredictability underscores a lack of robust and stable visual feature extraction and interpretation, contrasting with human visual processing, which tends towards predictable errors based on similarity.

\section{Usage of Generative AI tools}
We utilized Generative AI tools to help improve the language, phrasing, and readability of this manuscript.

\end{document}